\documentclass[sigconf,10pt]{acmart}

\setcopyright{acmlicensed}\acmConference[SoCC '24]{ACM Symposium on Cloud Computing}{November 20--22, 2024}{Redmond, WA, USA}

\renewcommand\footnotetextcopyrightpermission[1]{}

\usepackage[normalem]{ulem}
\usepackage{graphicx}
\usepackage{balance}
\usepackage{diagbox}
\usepackage{url}
\usepackage[utf8]{inputenc}
\usepackage{amsmath}
\usepackage{amsthm}
\usepackage{thmtools, thm-restate}
\usepackage{xspace}
\usepackage{epstopdf}
\usepackage{booktabs}
\usepackage{tabularx}
\usepackage{epsfig}
\usepackage[normalem]{ulem}
\usepackage{hyperref}
\usepackage[font=small]{subfig}
\usepackage{float}
\usepackage{cleveref}
\usepackage{xcolor,colortbl}
\usepackage{lipsum}
\usepackage{capt-of}
\usepackage{algorithm}
\usepackage[noend]{algpseudocode}
\usepackage{listings}
\usepackage{multicol}
\usepackage{multirow}
\usepackage[normalem]{ulem}
\usepackage{soul}
\usepackage{relsize}
\usepackage{stmaryrd}
\usepackage{wrapfig}
\usepackage{array}
\usepackage{enumitem}
\usepackage{verbatim}
\usepackage{xurl}
\usepackage{adjustbox}
\usepackage{setspace}
\usepackage{makecell} 

\usepackage{listings} % For better control over code formatting
\usepackage{xcolor} % For coloring the code

\definecolor{keywordcolor}{rgb}{0.13, 0.13, 1}
\definecolor{commentcolor}{rgb}{0.5, 0.5, 0.5}
\definecolor{stringcolor}{rgb}{0.58, 0, 0.82}
\definecolor{backcolor}{rgb}{0.95, 0.95, 0.92}

\newcommand{\rom}[1]{\uppercase\expandafter{\romannumeral #1\relax}}
\newcolumntype{L}[1]{>{\raggedright\let\newline\\\arraybackslash\hspace{0pt}}m{#1}}

\newcommand{\system}{\textsc{Hapi}}

\newcommand{\objstore}{\textsc{COS}}

\begin{document}

\title{Accelerating Transfer Learning with Near-Data Computation on Cloud Object Stores}

\author{Diana Petrescu}
\orcid{0009-0006-2229-235X}
\affiliation{%
  \institution{EPFL}
  \city{Lausanne}
  \country{CH}
}
\email{diana.petrescu@epfl.ch}

\author{Arsany Guirguis}
\orcid{0000-0002-0898-0387}
\affiliation{%
  \institution{EPFL}
  \city{Lausanne}
  \country{CH}
}
\email{arsany.guirguis91@gmail.com}

\author{Do Le Quoc}
\orcid{0000-0002-1433-0217}
\affiliation{%
  \institution{Huawei Munich Research Center}
  \city{Munich}
  \country{DE}
}
\email{quoc.do.le@huawei.com}

\author{Javier Picorel}
\orcid{0009-0009-6984-1303}
\affiliation{%
  \institution{Huawei Munich Research Center}
  \city{Munich}
  \country{DE}
}
\email{javier.picorel@huawei.com}

\author{Rachid Guerraoui}
\orcid{0000-0002-4794-8902}
\affiliation{%
  \institution{EPFL}
  \city{Lausanne}
  \country{CH}
}
\email{rachid.guerraoui@epfl.ch}

\author{Florin Dinu}
\orcid{0009-0005-1514-2997}
\affiliation{%
  \institution{Huawei Munich Research Center}
  \city{Munich}
  \country{DE}
}
\email{florin.dinu@huawei.com}

\begin{abstract}

Storage disaggregation underlies today’s cloud and is naturally complemented by pushing down some computation to storage, thus mitigating the potential network bottleneck between the storage and compute tiers. We show how ML training benefits from storage pushdowns by focusing on transfer learning (TL), the widespread technique that democratizes ML by reusing existing knowledge on related tasks. We propose HAPI, a new TL processing system centered around two complementary techniques that address challenges introduced by disaggregation. First, applications must carefully balance execution across tiers for performance. HAPI judiciously splits the TL computation during the feature extraction phase yielding pushdowns that not only improve network time but also improve total TL training time by overlapping the execution of consecutive training iterations across tiers. Second, operators want resource efficiency from the storage-side computational resources. HAPI employs storage-side batch size adaptation allowing increased storage-side pushdown concurrency without affecting training accuracy. HAPI yields up to 2.5$\times$ training speed-up while choosing in 86.8\% of cases the best performing split point or one that is at most 5\% off from the best.

%HAPI shows up to 2.5x training speed-up while splitting optimally or up to 5\% off in 86.8\% of cases.
\end{abstract}

\maketitle
\section{Introduction}

Storage disaggregation (i.e., the separation of the storage and compute tiers) powers today's cloud object stores (COS) (e.g. Amazon S3~\cite{amazon-s3}, Google Cloud Storage~\cite{google-cloud-storage}, Azure Blob Storage~\cite{sosp11-AzureStorage}) as it reduces costs and simplifies management by allowing the two tiers to scale independently.  Unfortunately, these benefits come at the cost of a potential network bottleneck~\cite{icde2020-pushdowndb, vldb2021-flexpushdowndb, amazon-aqua} between the tiers as network bandwidth growth is outpaced by storage bandwidth and compute throughput growth~\cite{nsdi21-alibaba-storage-rdma, micro20-aquoman-mit}. Near-data computation techniques are the natural complement to storage disaggregation. These involve provisioning storage-side compute resources to run part of an application (called a \textit{pushdown}) in order to mitigate the network bottleneck by reducing the amount of data transferred between tiers. These storage-side compute resources are limited by design as they are not meant to replace the compute tier. Following the initial success of pushdowns for a restricted set of workloads (e.g., SQL~\cite{amazon-aws-s3-select-blog}), there is renewed interest in broadening the applicability of such pushdowns to new applications and to specialized hardware.

This paper shows how ML training can benefit from pushdowns to disaggregated storage by focusing on transfer learning (TL)~\cite{tl-survey2020}, a widespread ML technique~\cite{mckinsey-tl} that enables a generic model previously trained (pre-trained) on a large dataset to be efficiently customized (fine-tuned) for a related task. TL democratizes ML by lowering the entry bar, as fine-tuning existing models avoids the need for new, large datasets and the computational expense of training models from scratch. Thus, TL has become a cornerstone of modern cloud ML services~\cite{sagemaker-tl, aws-bedrock, azure-databricks-tl, google-vertex-ai}, enabling the use of pre-trained models and scalable fine-tuning capabilities across major platforms.
In traditional TL fine-tuning, the initial DNN layers perform feature extraction while the rest perform re-training.

This paper proposes \system{}\footnote{HAPI was the Egyptian god of the annual flooding of the Nile, often portrayed as binding two regions (splits) of Egypt (\url{https://en.wikipedia.org/wiki/Hapi_(Nile_god)}).}, a new TL fine-tuning system that spans the storage and compute tiers and judiciously pushes down to storage part of the TL DNN. \system{} leverages two new techniques that address challenges introduced by storage disaggregation for the benefit of both users and operators. The first challenge is that pushdowns make it harder for applications and users to optimize performance. Typically, pushdowns are chosen to minimize network time by having the pushdown's output be smaller than the job's input. \system{} builds on the insight that, for reducing TL fine-tuning time, pushing down only to minimize network time is useful but, unfortunately, sub-optimal. Instead, applications need to carefully balance the pushdown processing time, the network transfer time as well as the compute tier processing time. \system{} achieves this balance by splitting the TL DNN during its feature extraction phase, which contains some DNN layers with relatively small output sizes (for reducing network time) (\S\ref{sec:motivation}), while also allowing the pushdown processing time for iteration $N+1$ to substantially overlap with the compute tier processing time for iteration $N$.

The second challenge, particularly important to operators, consists in using the limited storage-side compute resources efficiently. \system{} addresses this challenge with our novel technique called storage-side batch size adaptation. Splitting the TL DNN naturally decouples the batch sizes used in each tier. The insight is that one can use a significantly smaller batch size in the storage-side pushdown compared to the compute tier, importantly, without affecting training accuracy. By design, \system{} maintains the same training accuracy as if the training was fully performed in the compute tier (i.e., no pushdowns to storage) by pushing down only parts of the TL feature extraction phase which uses frozen weights (fixed, not re-trained) and by ensuring that the size and contents of the training batch in the compute tier remain unchanged. The benefit of storage-side batch size adaptation is that it greatly reduces the amount of GPU memory used by pushdowns as the initial layers of feature extraction are the most memory intensive (\S\ref{sec:motivation}) due to their larger output sizes. This memory reduction has two crucial consequences. First, this enables several pushdowns to make progress concurrently in the COS, thus improving resource efficiency. Second, this avoids the out-of-memory (OOM) errors that plague practitioners.

\system{} spans both the \objstore{} and compute tiers, is transparent to the user, and relies on inexpensive profiling runs. \system{}'s design uses a stateless storage-side component alongside lightweight request between the two tiers to simplify practical concerns regarding load balancing, scalability and failure resilience. 

Specifically, the contributions of this paper are:
\begin{enumerate}
    \item Identifying and demonstrating the benefits of applying near-data computation techniques to TL on top of the disaggregated \objstore{}.
    
    \item A measurement study of DNN layer characteristics across 7 popular DNNs (\S\ref{sec:motivation}), including modern architectures like the Vision Transformer (ViT) and widely-used CNNs such as ResNet and DenseNet. These characteristics play an important role in our system design.
    
    \item \system{}, an end-to-end system comprising two key design techniques: DNN splitting (\S\ref{subsec:splitting}) and storage-side batch size adaptation (\S\ref{subsec:ba-adapt}) which reduce network transfers, improve training runtime and enable increased pushdown concurrency in the \objstore{}. 
    
    \item An extensive evaluation (\S\ref{sec:eval}) showing up to 2.5$\times$ speed-up in training runtime while choosing in 86.8\% of cases the best performing split point or one that is at most 5\% off from the best.

\end{enumerate}

The paper is organized as follows. We provide background in \S\ref{sec:background} and present our measurement study in \S\ref{sec:motivation}. \S\ref{sec:design} presents the design and implementation of \system{}. We show experimental results in \S\ref{sec:eval} and discuss related work in \S\ref{sec:related-work}.

\section{Background} 
\label{sec:background}

\subsection{Transfer learning}
\label{background:tl}

TL democratizes ML by allowing knowledge from a model pre-trained on a large dataset to be adapted and reused (fine-tuned) for a different but related task~\cite{weiss2016survey}. In doing so, the DNN training time and the generalization error are reduced~\cite{deeplearning-book}. As models grow in size and complexity, TL has become increasingly essential for efficiently adapting pre-trained models to specific tasks with fewer resources, making their deployment more practical across various domains. The intuition behind TL is that the pre-trained model, often referred to as a backbone, captures generalizable embeddings (representations of the input data) which can be adapted or fine-tuned for new tasks rather than requiring a model to be trained from scratch. Examples of backbones include convolutional neural networks (CNNs) and Vision Transformers (ViTs) for computer vision~\cite{vaswani2017attention}, as well as models like BERT~\cite{devlin-etal-2019-bert} and GPT~\cite{NEURIPS2020_1457c0d6} for natural language processing (NLP). Recent advances in Vision Transformers~\cite{dehghani2023scalingvisiontransformers22, Grainger2022PaCaViTLP, wang2023crossformerversatilevisiontransformer} exemplify this approach by providing scalable and reusable backbones that can be fine-tuned for a wide range of tasks, including classification, detection, and segmentation.

%These advancements reflect the current trend in vision models towards creating versatile architectures that improve efficiency and flexibility in modern computer vision.

\begin{figure}[h!]
    \centering
    \includegraphics[width=0.8\linewidth]{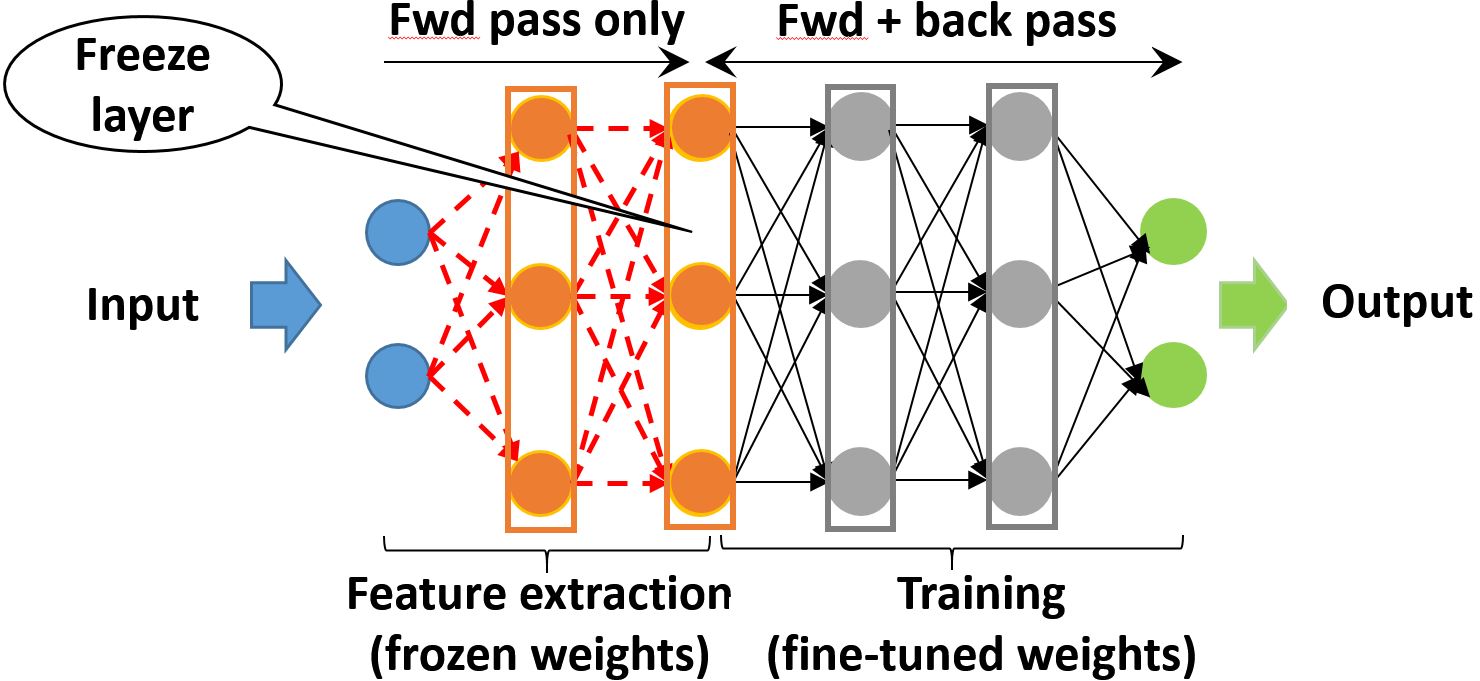}
    \caption{Overview of TL fine-tuning.}
    %\vspace{-2mm}
    \label{fig:finetuning}
\end{figure}

The key to TL lies in fine-tuning the pre-trained model, as depicted in Figure~\ref{fig:finetuning}. Pre-training is usually done on a different system and often by a different entity and is outside the scope of this paper. Traditionally, fine-tuning is divided further into two phases: {\em (i) feature extraction}, where embeddings or high-level representations are extracted from new input data using (partially or entirely) the pre-trained model, and then {\em (ii) training}, to create a new classifier using the extracted embeddings~\cite{deeplearning-book}. Typically, the early layers of the pre-trained model, which capture more general features (e.g., edges, textures in vision models), are frozen during fine-tuning, i.e., the model weights of these layers (in red in Figure~\ref{fig:finetuning}) are not updated with backpropagation. Every iteration (i.e., input batch processed) involves feature extraction followed by training. We refer to the last frozen layer as the \emph{freeze layer (or index)}.

\begin{figure*}[h]
\centering
\subfloat[Per-layer output sizes]{\includegraphics[width=0.33\linewidth,keepaspectratio]{figures2024/pdf-motiv-layer-out-size.pdf}
\label{fig:motiv-layer-output-sizes}}
\subfloat[Per-layer forward pass time]{\includegraphics[width=0.33\linewidth,keepaspectratio]{figures2024/pdf-motiv-fwd-pass.pdf}
\label{fig:motiv-fwd-pass-times}}
\subfloat[Per-layer max GPU memory usage]{\includegraphics[width=0.33\linewidth,keepaspectratio]{figures2024/pdf-motiv-layer-max-mem.pdf}
\label{fig:motiv-max-mem-usage}}
\caption{A measurement study of 3 per-layer properties for  7 popular DNNs.}
\end{figure*}

%\vspace{-1mm}
\subsection{Cloud object stores}
\label{background:cos}
%\vspace{-2mm}
Cloud object stores (\objstore{}), such as Amazon S3~\cite{amazon-s3}, Google Cloud Storage~\cite{google-cloud-storage}, and Azure Blob Storage~\cite{sosp11-AzureStorage}, are a popular way to store large-scale unstructured data, providing ease of use, high availability, high scalability, and durability at a low cost~\cite{object-storage1}. \objstore{} are the prime example of storage disaggregation. The \objstore{} is connected to the compute tier by a network that, unfortunately, is a bottleneck~\cite{object-storage2, object-storage3, nsdi19-fastslow, nsdi21-alibaba-storage-rdma} even when the network is maxed out. The reason are hardware trends. The network bandwidth between the \objstore{} and the compute tier is lower than the internal storage bandwidth of \objstore{} servers and also lower than the computation throughput~\cite{nsdi21-alibaba-storage-rdma, li2016hippogriffdb}. The typical network bandwidth of a single cloud server is 25 - 400 Gbps (3.125 - 50 GBps)~\cite{huawei-ecs,sigcomm21-huawei-acc,nsdi21-alibaba-storage-rdma, sigcomm24-alibabahpn}. A single modern NVMe SSD can read sequentially at well over 10 GBps~\cite{crucial-t705-ssd, sigmod25-nvme-arrays}, so a couple of NVMe SSDs are sufficient to max out the network bandwidth. In practice, storage servers are provisioned with many SSDs; an arrray of PCIe 5.0 NVMe SSDs can exceed 100 GBps in read throughput~\cite{sigmod25-nvme-arrays}. Other storage media are faster than SSDs, further aggravating the network bottleneck. With sufficient thread parallelism, DRAM reads can exceed 100 GBps, and persistent memory reads can exceed 30 GBps~\cite{fast20-swanson-pm}. Compute throughput also exceeds network bandwidth~\cite{li2016hippogriffdb, nsdi21-alibaba-storage-rdma}. Earlier studies~\cite{object-storage3,nsdi19-fastslow} have reported network read throughput as low as 100 MBps per connection from Amazon S3 but the trend is for network bandwidth to improve. More recently, up to 100 Gbps from general purpose instances to S3 has been reported~\cite{vldb23-obj-store-neumann}. 

In light of this enduring network bottleneck, an important trend has been to push down computation inside the \objstore{}, to reduce the amount of data sent over the network. Pushdowns were initially restricted to a subset of SQL (e.g., Amazon S3 Select~\cite{amazon-aws-s3-select-blog}), but there is a renewed effort in the industry to support more complex pushdowns for computations such as image processing~\cite{alibaba-oss-image-proc} or analytics~\cite{amazon-aqua}. There has been growing interest in pushing down parts of ML computations to storage~\cite{sophon-hotstorage24, ndpipe-asplos24}. This trend goes hand-in-hand with another development, enabling pushdowns to use specialized hardware (\S\ref{background:hardware}).

Despite these trends, two challenges remain for \objstore{} users. The network may remain a bottleneck despite the pushdowns, and the \objstore{} computational resources are scarce and need to be used efficiently as they are only meant to mitigate the network bottleneck and not replace the compute tier.

%\vspace{-1mm}
\subsection{Hardware-accelerated pushdowns}
\label{background:hardware}
%\vspace{-2mm}
Pushdowns were initially restricted to a subset of SQL, including filtering, projecting, and aggregation (e.g., Amazon S3 Select~\cite{amazon-aws-s3-select-blog}). 
The current, natural trend is to offer the benefits of pushing down to a wider range of applications. Unfortunately, restricting pushdowns to CPUs can lead to wasted resources and performance. First, for more complex operations, CPUs can become a bottleneck. Studies show that even with 32 cores, an SGD optimizer can fully utilize the CPU when using a 100 Gbps network~\cite{osdi20-bytePS}. Second, it is not sufficient for the CPU processing to be just faster than the network because the output of a pushdown may be smaller than its input. For example, for a pushdown to generate an output at 100 Gbps, assuming an input/output ratio of 2, it needs to process input at 25 GBps. Finally, the aggregate storage bandwidth of a storage server tends to increase faster than CPU capabilities~\cite{micro20-aquoman-mit}. 

As a result, the current trend is to allow pushdowns to use specialized hardware such as GPUs. Several works~\cite{chang2020ecs2,haddock2017gpu} have proposed to use them in storage systems to speed up erasure coding. Finally, there is a push to more-closely integrate storage with GPUs, which further increases the appeal of next-to-storage GPUs. For example, IBM's Storage Scale System 6000~\cite{ibm-storage-scale-system} integrates NVIDIA GPUDirect Storage~\cite{nvidia-gpudirect-storage}, enabling a direct data path between GPU memory and storage to reduce latency and enhance performance for AI and data-intensive workloads.

%IBM is offering high-performance storage with NVIDIA GPUs~\cite{ibm-nvidia,ibm-objectstore-ai}

%Finally, there is a push to more-closely integrate storage with GPUs~\cite{gpudirect, nvidia-swiftstack}, which further increases the appeal of next-to-storage GPUs.

%In this work, we show that the feature extraction phase can be pushed down to the \objstore{} tier to reduce the training latency and the network traffic between the compute tier and the \objstore{} tier while also maximizing the utilization of the \objstore{} computing resources such as the GPU memory.
\section{Measurement study}
\label{sec:motivation}

Next, we present a detailed measurement study of 7 DNNs. These include a state-of-the-art Vision Transformer~\cite{vaswani2017attention} as well as several widely-adopted foundational models such as ResNet50~\cite{he2016deep}, DenseNet121~\cite{huang2017densely}, and VGG19~\cite{simonyan2014very}. These models cover a diverse range of architectural characteristics, making them well-suited for evaluating system-level performance in terms of speed and resource efficiency. We characterize the per-layer properties across three dimensions: output size, compute time, and maximum GPU memory used. These properties all play a role in \system{}'s design (\S\ref{sec:design}). Additional layer-related information for each DNN can be found in Table~\ref{table:models}. For the DNNs structured as a sequence of blocks (e.g. ResNets) we count one block as one layer and we use only block boundaries as candidate split points. The input dataset is ImageNet. For readability purposes we group the models into 3 sub-groups in each figure.

\vspace{0.05in}\noindent{\bf Hardware setup. }
For this section we use two identical GPU-accelerated machines from a public cloud one for the \system{} client and the other for the \objstore{} and the \system{} server. Each machine has an Intel Xeon Gold 6278C CPU with 16 cores, 64 GB RAM, 1 Tesla T4 GPU with 16 GB RAM, a 300 GB SSD and runs Ubuntu 20.04 server 64 bit with CUDA 11.2. The network bandwidth between VMs is 12 Gbps. 

\vspace{0.05in}\noindent{\bf Per-layer output size. }
Figure \ref{fig:motiv-layer-output-sizes} shows the per-layer output size. For reference, the size of a pre-processed ImageNet input tensor is shown with the horizontal dotted blue line. This output size is for a batch size of 1. One can accurately extrapolate from this by multiplying by a specific batch size. 

The important observation is that for most models, early on in the DNN structure, there exist layers with a comparatively small output size, often significantly smaller than subsequent layers (e.g. ResNet layer 4, AlexNet layers 3 and 6). These layers are good candidates for splitting the TL application between the \objstore{} and the compute tier. Overall, the layer output size generally increases in the beginning (with convolution layers) and then decreases (with pooling layers) but not in a monotonic fashion. While only a few split points have an output size smaller than the pre-processed tensor, we will show that this is not strictly necessary for performance gains because optimizing the network time is only part of the story. \system{} owes its benefits to balancing network time against the computation time in both tiers.

\vspace{0.05in}\noindent{\bf Per-layer computation time. }
Figure~\ref{fig:motiv-fwd-pass-times} shows the per-layer computation time for the forward pass for a batch size of 128. The insights are that some models are more lightweight (e.g. AlexNet) than others and that some models (e.g. DenseNet) show significant variability across layers. However, the forward pass time remains relatively stable with an increase in layer index and gives \system{} the flexibility to balance computation time across the two tiers.

\vspace{0.05in}\noindent{\bf Per-layer maximum GPU memory usage. }
Figure~\ref{fig:motiv-max-mem-usage} shows the maximum amount of GPU memory needed for the processing of each layer for a batch size of 128. Again, the variation is model dependent and for each model there is variability between layers. Generally, the first few layers use more memory suggesting that these need to be the focal point if reducing memory consumption is desired.

\vspace{0.07in}\noindent{\bf The need for a dynamic splitting algorithm. }
We next motivate \system{}'s dynamic splitting approach by showing the limitations of static splitting approaches. 

Figure~\ref{fig:motiv-need-splitting} shows 7 groups of bars. Each group represents a model (listed on the x-axis) and the 7 bars in each group represent different batch sizes (128, 256, 512, 1024, 1536, 2048, 3072 from left to right). The figure summarizes a performance sweep over all possible split points in the DNNs (after each layer). The figure shows per-epoch runtime normalized to splitting at the freezing layer (@Freeze). The points (blue rectangle, red circle) show two other sensible static splits: (1) with red circles @Min, splitting at the layer with the smallest output (earliest such layer as a tiebreaker) and (2) with blue rectangles NoSplit, i.e., sending the entire input to the compute tier. The lines show the speed-up range for all other splits, except the three mentioned above (@Freeze, @Min, NoSplit). If some of the static splits do not appear (e.g. Vgg11, Vgg19, Transformer) it is because they cause OOMs on the client. However, @Freeze (y-axis value = 1) never OOMs. 

\begin{figure}[h]
    \centering
    \includegraphics[width=\linewidth]{figures2024/pdf-motiv-split-choices2.pdf}
    \caption{Speed-up when splitting at various layers normalized to splitting at the freezing layer.}
    \label{fig:motiv-need-splitting}
\end{figure}

%%Keep in mind that this already does BA !

Several insights can be derived from the figure, which point to the need for an intelligent dynamic splitting algorithm. First, no static splitting strategy wins in all cases. Second, even though one splitting strategy is the best option in some cases (e.g. @Min for ResNet18), it can be very sub-optimal in other cases (e.g. @Min for DenseNet or Vgg19). Third, the behavior is model dependent and the training batch size also matters. Finally, some splits lead to OOMs on the client side and thus should be avoided. This is noticeable in the fact that some splits do not show up in the figure (e.g. NoSplit for larger batch sizes for several models or @Min for Transformer). For the largest 3 batch sizes for Transformer only @Freeze does not result in an OOM which results in no vertical bars (i.e., the speed-up range is 1).

\section{Design and Implementation}
\label{sec:design}

% Notes: you use interchangeably COS and storage-side.

\subsection{Main insights}

HAPI's two main techniques (1) splitting the TL DNN between the storage and compute tiers and (2) storage-side batch size adaptation are based on the following insights:
\begin{enumerate}
    \item Judicious splitting of the DNN's feature extraction phase between the storage and compute tiers can minimize training time by balancing two factors: the amount of data sent over the network and the overlap in processing time on the storage and compute tiers (\S\ref{subsec:splitting}).
    
    \item Splitting the DNN naturally decouples the batch size of feature extraction from that of training and thus allows \system{} to adapt the former (\S\ref{subsec:ba-adapt}). This yields significant reductions in the storage-side GPU memory used, avoiding OOM errors and enabling increased pushdown concurrency in the \objstore{}.
    
    \item {\it Adapting the storage-side feature extraction batch size does not affect training accuracy}. By design, \system{} maintains the same training accuracy as if the training was fully performed in the compute tier without pushdowns and without batch size adaptation. The reasons are the following. First, \system{} keeps the training batch size and its contents unaffected. Second, feature extraction uses frozen weighs (fixed, not re-trained) so its output is deterministic.

\end{enumerate}

\subsection{Design overview}

\begin{figure}[t]
\centering
\includegraphics[width=0.42\textwidth]{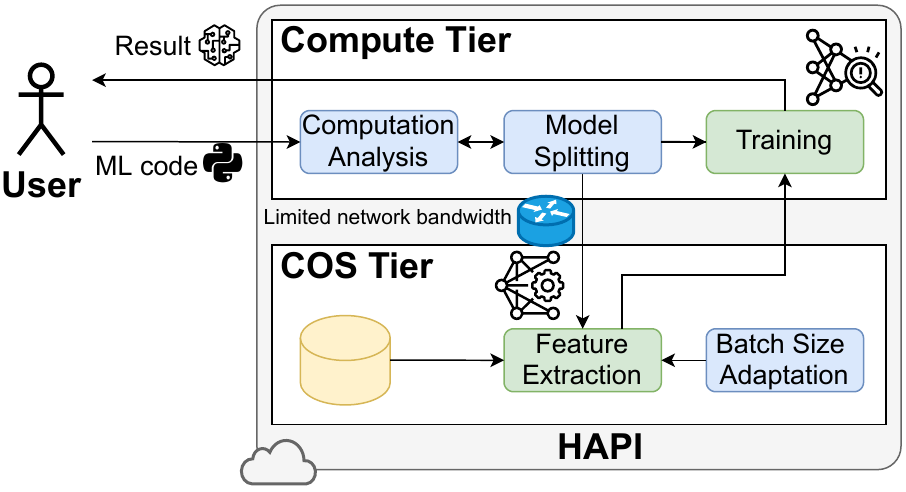}
%%\vspace{-2mm}
\caption{The high-level architecture of \system{}. The \system{} client resides in the compute tier, and the \system{} server resides in the \objstore{}.}
\label{fig:system-overview}
\end{figure}

\vspace{0.05in}\noindent{\bf At the high level. }
As shown in Figure~\ref{fig:system-overview}, \system{} has two components: the \system{} client, which runs on the compute tier, and the \system{} server, which runs in the \objstore{}. The client decides at which layer to split the DNN using a runtime estimation model which leverages in-job profiling information. The server executes feature extraction up to the layer decided by the client while employing batch size adaptation. Each storage machine runs one \system{} server.

The communication between the \system{} clients and servers occurs at the granularity of a {\it request}. These requests are lightweight and several may be sent by the client to obtain data for a single training iteration especially if the training batch size is large. This is key to leveraging parallelism in the storage tier as one or several \system{} servers can process requests from one iteration in parallel. \system{} servers treat requests independently, in a stateless manner. This combination of lightweight requests and stateless storage-side processing has practical benefits: it simplifies load balancing, failure recovery and scalability as any storage server can handle any request.

On the \objstore{} side we differentiate between proxy and storage nodes. The \system{} server runs on the \objstore{} proxy. There is a fast network between the \objstore{} proxy and the \objstore{} storage nodes which host the DNNs and the dataset. However, the network between the compute tier and the \objstore{} is more limited in bandwidth.

%This is based on our object store service which employs this two layer design.

\vspace{0.05in}\noindent{\bf Terminology. }
The DNN layer at which \system{} decides to split the DNN is the \emph{split index (or layer)}. The \emph{freeze index (or layer)} is the DNN layer that separates feature extraction from training, and it is chosen by the user. The \emph{training batch size} is the batch size specified by the user, and it is used in the training phase. \system{} may decide to use a different batch size for computation on the \objstore{} (\S\ref{subsec:ba-adapt}); we call this the \emph{\objstore{} batch size}.
We use \emph{POST request} to refer to a request sent from the \system{} client to the \system{} server. We use \emph{storage request} to refer to a request sent by the \system{} server to the storage nodes.

\begin{figure}[t]
\centering
\includegraphics[width=0.48\textwidth]{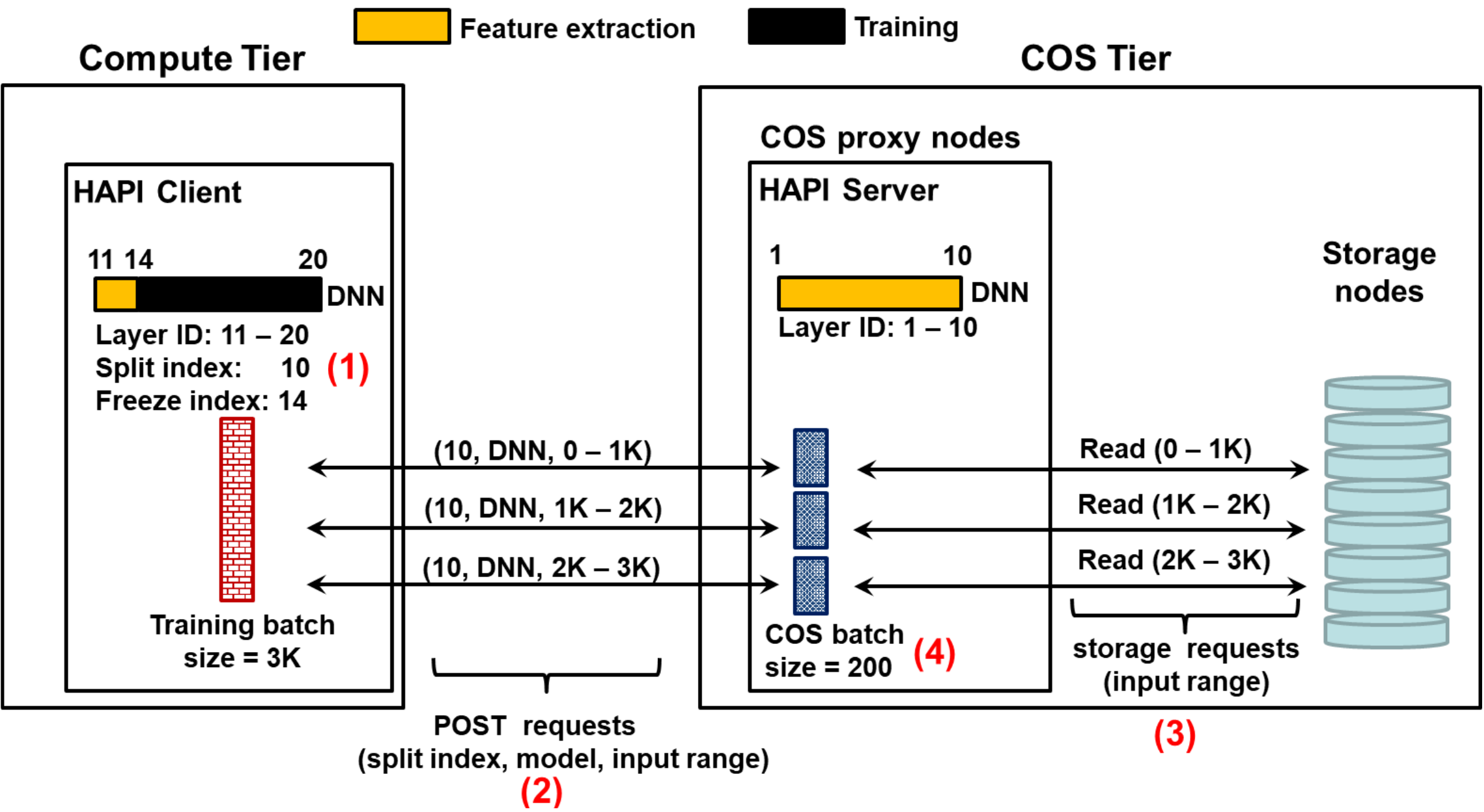}
%%\vspace{-4mm}
\caption{\system{} request flow. A TL job is split at layer 10 with training batch size 3000. 3 requests are sent in parallel to the \system{} server. On the \objstore{}, for each request, feature extraction uses a smaller \objstore{} batch size of 200.
}
\label{fig:request-flow}
\end{figure}

\vspace{0.05in}\noindent{\bf Request flow.}
To initiate a TL computation, users provide their application to the \system{} client, which extracts the application configuration (e.g., the model type, the dataset, the freeze index) and, after some profiling, splits (\S\ref{subsec:splitting}) the DNN into 2 parts: one to be executed on the \objstore{} and the other to be executed on the compute tier. Figure~\ref{fig:request-flow} depicts an example request flow between the two tiers during one iteration, with a training batch size of 3000 and a split index of 10. The values in Figure~\ref{fig:request-flow} are for illustration only.

For each training iteration, the \system{} client sends to the \system{} server HTTP POST requests containing the necessary information: split index, model type, and the name of the object that stores the corresponding data batch. Several such POST requests may be sent during one iteration to benefit from parallelism on the storage side. In Figure \ref{fig:request-flow}, the client sends 3 requests to the storage, each for 1000 objects. Together, these responses for these 3 requests will provide all data for the training batch size of 3000.

When the \system{} server receives a request, it first reads in the training data and the model by sending a \emph{storage request} to the \objstore{} storage nodes and then executes the feature extraction part up to the split index. As an optimization, the server may reuse a model if it already exists in its GPU memory (e.g., from previous requests in the same training job) because weights are frozen on the storage side so requests do not modify the model. Thus, such model reuse does not go against the stateless nature of \system{}'s \objstore{} processing.

The \system{} server chooses at will the \objstore{} batch size to be used for feature extraction (\S\ref{subsec:ba-adapt}). In Figure \ref{fig:request-flow}, the server uses a \objstore{} batch size of 200 for each request which allows it to execute all 3 requests in parallel without encountering OOMs. That is, the server further divides a request's work for 1000 objects into 5 sequential partitions of 200 to reduce the memory requirements.

After finishing the feature extraction portion assigned to it, the \system{} server sends back the outputs of the split index to the \system{} client. The client combines the outputs for all requests sent during that iteration and continues the TL computation on the compute tier. Note that the client uses the training batch size for its entire computation even while executing the last part of feature extraction (if any).

\vspace{0.05in}\noindent{\bf Observations.}
First, \system{} only executes feature extraction next to storage in the \objstore{}. Consequently, this means that the split index will always be smaller than or equal to the freeze index (i.e., the split is always before the training phase). 

Second, as it is typical in current ML frameworks, HAPI pipelines both communication and computation on the client side. That is, requests for iteration $N + 1$ are already sent to the storage just before the client starts its processing for iteration $N$.

Third, scaling down the \objstore{} batch size does not affect the convergence and accuracy of TL. This is because the \objstore{} batch size is used solely for feature extraction rather than for the actual training.

Fourth, for the requests belonging to one iteration, it is possible for the client to receive results out of order (i.e., the first request sent may not complete first) as the request completion time depends on the \objstore{}. To address this, the \system{} client re-orders the intermediate results returned from the server to ensure that the order of data in a training batch does not change, thus maintaining the learning trajectory.

\subsection{Splitting TL between storage and compute tiers}
\label{subsec:splitting}

Splitting the TL computation between the storage and compute tiers is the cornerstone of \system{}'s design. It enables training runtime improvements by navigating the interplay between the amount of data sent over the network and the overlap in processing time between the client and storage sides. Furthermore, it decouples the batch sizes used on the client and storage sides, paving the way for benefits via storage-side batch size adaptation (\S\ref{subsec:ba-adapt}).

\system{}'s approach to splitting leverages several components which are detailed in this subsection: (1) a fine-grained profiler, (2) a model that uses the profiler, (3) a GPU memory usage estimator and most importantly (4) a splitting algorithm which uses all other components. HAPI's contributions are (1), (2) and (4) while for (3) it leverages existing techniques.

\vspace{0.05in}\noindent{\bf The splitting algorithm.} {\it The goal of the splitting algorithm is to minimize the training runtime for a job by choosing the proper splitting point.}

\begin{figure}[t]
\centering
\includegraphics[width=0.48\textwidth]{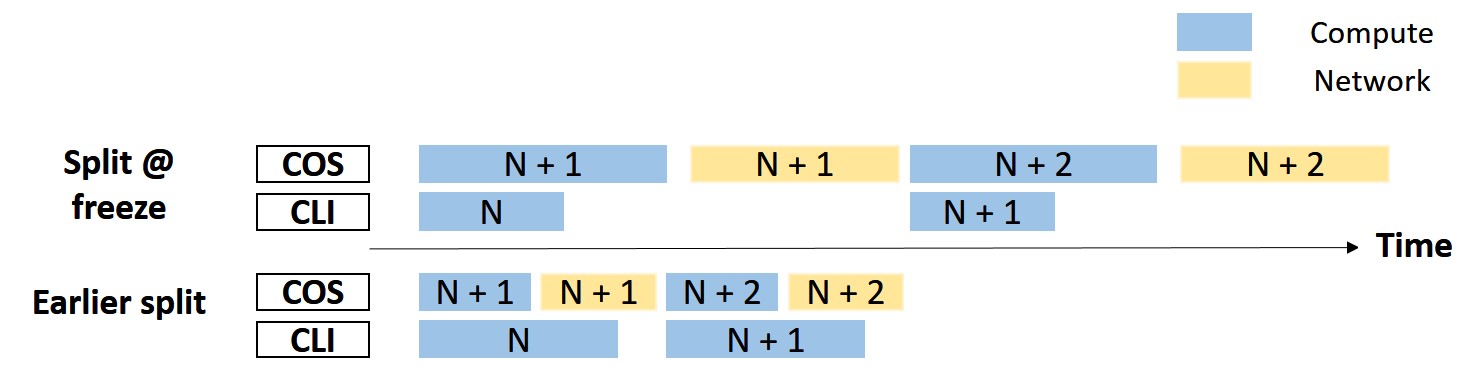}
\caption{Adjusting the overlap between the \objstore{} and the client by adjusting the split index.}
\label{fig:overlap}
\end{figure}

Central to \system{}'s splitting algorithm is the concept of overlap between the processing on the client and storage sides and its interplay with the chosen split point. This is illustrated in Figure~\ref{fig:overlap}. During training, \system{} overlaps the storage-side processing for iteration $N + 1$ with the client side processing for iteration $N$. The dominating side ends up dictating the training runtime. By adjusting the split index, \system{} can adjust the amount of overlap and thus directly influence the training runtime. One can think of the overlap as saved runtime because both the storage and client tiers make progress concurrently instead of sequentially. Thus, intuitively, the right strategy is to maximize the overlap. Overall, \system{} balances both network and computation time in the decision process which is essential in minimizing training runtime. 

When choosing the split point, accounting only for network time or also accounting for computation time but without accounting for the overlap are both sub-optimal. For example, the motivation section (\S\ref{sec:motivation}) showed that splitting at the layer with the smallest output is far from optimal. Intuitively, minimizing the network cost alone is insufficient because this is often achieved at a split point that pushes too much work on the storage side, reducing the overlap with the client side. One can also consider both network and computation times but not the overlap by minimizing the sum of the client time, server time and network time. This is a sensible approach inspired by ML inference~\cite{kang2017neurosurgeon} where for any single inference request there are no iterations and thus no overlap. In our evaluation (\S\ref{sec:eval}) we show that this solution also under-performs. The larger the overlap, the more this solution will over-estimate training runtime resulting in sub-optimal split choices.

\begin{algorithm}
    \footnotesize
	\caption{Choose the split index - runs on client}
	\label{alg:choose-split}
	\begin{algorithmic}[1]
    \If {Right before 1\textsuperscript{st} iteration of epoch 0}
        \State split\_idx = freeze\_idx
        \State collect profiling data from server \& client
        \If{job failed with OOM} 
            \State return error to client
        \EndIf
    \EndIf
    \item[]
    \If {Right before mid iteration of epoch 0}
        \State split\_idx = earliest split not predicted to cause OOM on client
        \State collect profiling data from server \& client
    \EndIf
    \item[]
    \If {At end of epoch 0}
        \State {runtimes=\{ \}}
        \For {potential\_split in $[0, freeze\_idx]$}
        \If {potential\_split predicted to NOT cause OOM on client}
            \State {derive RT\_SRV at potential\_split from profiling \& modeling}
            \State {derive RT\_CLI at potential\_split from profiling \& modeling}
            \State {derive RT\_NW at potential\_split from profiling \& modeling}
            \State {NR\_IT = nr iterations in an epoch}
            \State {runtimes$[potential\_split]$ = 
            \State RT\_SRV + max(RT\_SRV + RT\_NW , RT\_CLI) * (NR\_IT - 1) + RT\_CLI}
        \EndIf
        \EndFor
        \State split\_idx=idx such that runtimes[idx] is minimized
    \EndIf        
    
	\end{algorithmic}
\end{algorithm}

Algorithm~\ref{alg:choose-split} presents the steps involved in choosing the split index. This algorithm runs on the client. By default, \system{} uses the first epoch for profiling on both the server and client sides and then uses that information to select a split index for the rest of the epochs. While not default in \system{}, if desired, another epoch can be easily turned into a profiling epoch later on. Usually many epochs are needed for convergence so using one for profiling has negligible impact. Given a stable configuration (i.e., the same split), the behavior of ML training is deterministic across iterations and this ensures the relevance of the profiling.

For the first half of epoch 0 (Alg1:1-5), \system{} uses the freeze index as a split point. This is the minimum amount of work that can be done on the client. If this causes an OOM on the client, then the client is notified because any other split will also OOM. For the second half of epoch 0 (Alg1:6-8), the split index is changed to the earliest split that \system{} predicts will not cause an OOM on the client. Thus, this is the maximum amount of work that the client can perform. This earliest split could be 0 which means that there is actually no splitting (i.e. the input data is sent to the client which does the entire processing). At the end of epoch 0 (Alg1:9-19), \system{} selects the split for subsequent epochs. For each split that will not cause an OOM on the client, the algorithm estimates, using profiling and modeling information, the computation time on the server and client as well as the network time. Using these, it then estimates the per-epoch runtime while accounting for the overlap (Alg1:18). In particular, \system{} considers the overlap between (1) the client side and (2) the server side + network time because the client cannot start processing until it finishes receiving the data from the server.

A few more notes about the splitting algorithm. First, the algorithm is not concerned with the memory usage and OOM on the server side because it does not have access to that information. This is taken care of on the server side by \system{}'s batch size adaptation algorithm. Second, the algorithm may end up choosing to not split at all (split index 0) if it infers that to be the best solution. To estimate epoch runtime with split index 0, \system{} uses a simpler procedure than in Algorithm~\ref{alg:choose-split}. If split 0 is chosen as the final split, then it means it was already chosen as the earliest split for profiling in the second part of epoch 0 and thus per-iteration runtime is already available. Third, profiling using both the earliest and latest possible splits is particularly important for the case when the client and server have different GPU models with different performance. With a single split, profiling information from the server would have to be used for making estimates on the client side and vice-versa. This can be error prone with heterogeneous GPUs and is avoided with \system{}'s profiling method.

\vspace{0.05in}\noindent{\bf Profiling and modeling runtime.}
To derive epoch runtimes, as described above, we first break down the execution time on both client and server into fine-grained execution phases and estimate per-phase runtime using a combination of profiling and modeling.

Several of the phase costs are proportional to the split layer's output size so they are amenable to modeling. To obtain per-layer output sizes, there is a short profiling phase on the client (not included in Algorithm~\ref{alg:choose-split}), before the job starts, in which the client runs a forward pass over the DNN with a batch size of 1 with a synthetic input. From this, the layer output size for any potential split and batch size can be simply derived.

\begin{table}[h]
    \normalsize
    \def\arraystretch{1}
    \centering
    \footnotesize
    %\begin{tabular}{|l|l|l|}
    \begin{tabular}{|l|p{3cm}|p{2cm}|}
        \hline
        {\bf Profiling side} & {\bf Phase description} & {\bf How to estimate} \\\hline
        \hline
        @server & read data from storage & profiled\\\hline
        @server & read model to GPU & profiled\\\hline
        @server & move data to GPU & profiled\\\hline
        @server & forward pass & profiled\\\hline
        @server & move data out of GPU & linear w.r.t data size\\\hline
        @server & pickle for network transfer & linear w.r.t data size\\\hline
        \hline
        @client & network bandwidth & profiled \\\hline
        @client & unpickle received data & linear w.r.t data size\\\hline
        @client & create dataloader & linear w.r.t data size\\\hline
        @client & copy data to GPU & linear w.r.t data size\\\hline
        @client & forward pass & profiled \\\hline
        @client & backward pass & profiled \\\hline
    \end{tabular}
    \vspace{0.1in}
    \caption{Profiled execution phases including the location where the profiling takes place.}
    \label{tab:profiled-quantities}
\end{table}

Table~\ref{tab:profiled-quantities} shows the phases that we profile or model. The modeled phases have a cost that is proportional to the output size. The cost can be derived with knowledge of the layer output sizes and characteristics of the environment (e.g. memory bandwidth, pickle throughput). The environment characteristics are learned once per machine type and are job-agnostic.

The final step is to estimate the entire server-side and client-side processing times ($RT\_SRV$ and $RT\_CLI$ in Algorithm~\ref{alg:choose-split}) using per-phase estimates. The main challenge lies on the server-side in accounting for the impact of concurrency and waiting time. For concurrency, we model the GPU-only forward pass as sequential execution as our GPU (Nvidia T4) is time-shared. We consider all other costs perfectly parallelizable. Request waiting time can become a factor when some requests are queued at a storage server waiting for other requests to finish. For the case of concurrency among requests from the same training job, we modeled waiting time as the execution time of the wave of requests in front. To explain, consider 8 requests sent to a storage server of which 4 can be executed in parallel. The waiting time for the last 4 requests is the execution time of the first 4 requests. For the general cases of different concurrent jobs, the \system{} server would have to expose the expected waiting time to the \system{} client.

\vspace{0.05in}\noindent{\bf Memory usage estimation} A memory estimator is needed for Algorithm~\ref{alg:choose-split}. There is already past work on this topic~\cite{estimate-gpu-mem} and \system{} does not make a contribution in this respect. We use a simple estimation approach based on the intuition that the maximum amount of memory for the forward pass occurs at the layer which has the largest input + output size. We obtain these sizes as described above. We couple this approach with a past understanding of the jobs we run.

We note the following insights. Under-estimating memory for a specific split can can cause client-side OOMs. Over-estimating memory usage can cause, in some cases, skipping optimal split points. This can occur, for example, when the estimate points to a client-side OOM for the optimal split while in reality there is no OOM. Thus, over-estimation is preferable to under-estimation.

\subsection{Storage-side batch size adaptation}
\label{subsec:ba-adapt}

Batch size adaptation in this paper refers to \system{} reducing the \objstore{} batch size on the server side. As discussed, this is enabled by splitting which decouples the client and \objstore{} batch sizes, thereby allowing both to be chosen independently. As previously mentioned, batch size adaptation does not affect the final training accuracy. This is because batch size adaptation only applies to the feature extraction phase.

\vspace{0.07in}\noindent{\bf Memory saving. }
The main benefit of batch size adaptation comes from dramatically reducing the amount of GPU memory needed on the storage-side. A positive effect of that reduction is that this frees sufficient GPU memory to allow increased server-side request parallelism.

Without batch size adaptation, the GPU memory needed is a function of the {\it training batch size} and layer output sizes. If the user employs large training batch sizes then this runs the risk of OOMs. With batch size adaptation, the storage-side GPU memory needed is a function of the {\it \objstore{} batch size} and layer output sizes. There are two reasons why this results in significant server-side GPU memory savings. First, the \objstore{} batch size can be orders of magnitude smaller than the training batch size. Second, as shown in \S\ref{sec:motivation}, the layers with the largest output sizes (and which can consequently yield the largest memory savings) appear in the first part of the DNN which is exactly the part that is pushed down to storage and benefits from batch size adaptation.

Note that \system{}'s design based on requests is already a manifestation of batch size adaptation since the training batch on the client side is reconstructed from several smaller requests sent to the storage. \system{} applies batch size adaptation further inside a single request by reducing the \objstore{} batch size.

\vspace{0.07in}\noindent{\bf Choosing a \objstore{} batch size.} We found that using a static \objstore{} batch size across all workloads is sufficient because we found no relationship between \objstore{} batch size and a request's processing time on the storage side. We found no benefits from increasing the \objstore{} batch size beyond a certain value. The main thing to avoid is choosing a value that is too low which results in unnecessary overhead. \system{} uses a storage batch size of 16 objects.

An added benefit of this approach is that the batch size assignment is quick. More complicated approaches may delay the completion time especially for particularly lightweight requests. Indeed, we saw this initially when we experimented with a dynamic programming approach that chose a potentially different \objstore{} batch size for every request already queued at the server.

\subsection{System implementation}
\label{sec:impl}

We use PyTorch~\cite{paszke2019pytorch} for ML computation on both the client and server sides.

\vspace{0.07in}\noindent{\bf The \system{} server. }
We integrated \system{}'s server in the \objstore{} proxy server by incorporating the batch size adaptation algorithm and additional functions for ML computation. When the server receives a POST request, it reads the object holding the training data from storage and passes it to the ML function for processing. Each incoming request is served by a separate OS process. On the ML side, we created custom DNN models that are capable of running the forward pass between arbitrary start and end layers. This enables us to split the computation at any chosen layer. 
%\textcolor{blue}{\sout{Although we have not yet automated the customization of the models, we believe it is feasible and plan to do it as future work.}}

\vspace{0.07in}\noindent{\bf The \system{} client. }
Apart from choosing the split index for the model, the \system{} client is responsible for running the user's TL code. To integrate with our system, two modifications were necessary in the vanilla training code.
First, we use our custom models to enable starting computation from an arbitrary layer instead of being restricted to the default first layer.
Second, instead of streaming the raw training data using HTTP GET requests, we stream the intermediate outputs (resulting from the last layer executed on the \objstore{}) using HTTP POST requests.
Moreover, users provide the same training parameters in both cases, whether using the status quo or \system{}, and hence, the whole process is transparent to them.

\section{Evaluation}
\label{sec:eval}

\subsection{Methodology}
\label{subsec:eval_setup}

\noindent{\bf Applications. }
\system{} is application-agnostic: its benefits are orthogonal to the training objective, i.e., any TL application that trains a DNN can use \system{}. It can be used for speech recognition~\cite{eberhard2021effects,hannun2014deep} or language modelling~\cite{brown2020language}.
In our evaluation, we focus on image classification as our TL application for two reasons: (1)~due to its wide use in both academia and industry, and (2)~since the data points (i.e., images) used for such applications are large compared to those of other applications (e.g., text or audio files) and thus create a more challenging environment in terms of communication or computation bottlenecks.

We evaluate a Vision Transformer~\cite{vaswani2017attention} model along with popular, foundational DNNs from computer vision (ResNets, Vggs, etc). Table~\ref{table:models} lists all models and their freeze layers. The Transformer is a Vision Transformer from the PyTorch TorchVision library. Although originally designed for language modelling tasks, the Transformer has seen extensive applications beyond NLP to image classification~\cite{khan2021transformers}, video understanding~\cite{bertasius2021space}, and biological sequence analysis~\cite{nambiar2020transforming}. For the DNNs structured as a sequence of blocks (e.g. ResNets) we count one block as one layer and we split only at block boundaries. The mix of models from Table~\ref{table:models} allows us to cover a wide range of situations (compute intensiveness, output size variation, model size) that influence \system{}'s behavior.

\begin{table}[ht]
\small
\def\arraystretch{1}
\centering
\footnotesize
\begin{tabular}{|l|l|l|l|l|l|l|l|}
\hline
         {\bf MODEL} & {\bf Alex} & {\bf Res} & {\bf Res} & {\bf Dense} & {\bf Vgg} & {\bf Vgg} & {\bf Trans} \\
        {\bf } & {\bf Net} & {\bf Net18} & {\bf Net50} & {\bf Net} & {\bf 11} & {\bf 19} & {\bf former} \\\hline
Freeze & 17 & 11 & 21 & 20 & 25 & 36 & 17 \\ \hline
Total & 22 & 14 & 22 & 22 & 28 & 45 & 21 \\ \hline
\end{tabular}
\vspace{0.1in}
\caption{The 7 models used along with the index of the freezing layer and the total number of layers.} 
\label{table:models}
\end{table}

\vspace{0.05in}\noindent{\bf Metrics. }
As the main metric we use the time to finish one epoch of training. We present this either as relative speed-up when comparing two systems (e.g., by normalizing the runtime of a system to that of a competitor) or as absolute values. We use the runtime of epoch 1 as it is representative of all the epochs that follow. We do not show the training accuracy because, as discussed, \system{} keeps it unchanged by design. 

\vspace{0.05in}\noindent{\bf Training input. } 
We use the ImageNet~\cite{deng2009imagenet} dataset. As the actual input to the DNN we feed tensors resulting from separately pre-processing the input image. This is in line with current approaches to decouple input pre-processing from training ~\cite{isca22-dlrm-preproc-meta-kozyrakis} in order to not bottleneck the training. The input tensors are already stored on the proxy nodes in CPU DRAM. This allows us to focus on the communication between the client and the server. The tensor size is 224 x 224, the standard input size for vision tasks. We reuse the first 30720 ImageNet images for each training epoch as the pre-processed tensor is always the same size regardless of the input image. Nevertheless, \system{} can also take images as input and perform pre-processing right before training.

The POST request size is set to 128 objects. We use 7 training batch sizes (BS for brevity) that are multiples of the request size, ranging from small to very large: 128, 256, 512, 1024, 1536, 2048, 3072. Thus, the number of parallel requests sent by the \system{} client to the \objstore{} during one iteration equals BS divided by 128.

Batch size adaptation is by default applied to all requests on the server side and we set the default \objstore{} batch size to 16. The server sends the data to the client only after finishing the processing for all 128 objects in a request. On the server side, a maximum of 4 requests can be executed concurrently.

By default, \system{} does not reuse the DNN stored in GPU RAM between requests so it re-loads it into the GPU RAM for every new request. We evaluate the benefits of model reuse in \S\ref{subsec:model-reuse}.

\vspace{0.05in}\noindent{\bf \system{} configuration. } 
We use cloud VMs, one for the storage and one for the compute tier. Each VM has 64 GB RAM, 16 Xeon Gold 6278C vCPUs and one 16 GB Nvidia T4 GPU and runs Ubuntu 20.04 server 64 bit with CUDA 11.2. The network bandwidth between VMs is 12 Gbps. This is larger than the reported network bandwidth between client tiers and object stores in the past~\cite{object-storage3, nsdi19-fastslow} and reflects the current trend towards increased network bandwidth in data centers~\cite{vldb23-obj-store-neumann}. Nevertheless, we analyze the sensitivity to network bandwidth in \S\ref{subsec:vary-nw-bw}.

\vspace{0.05in}\noindent{\bf Competitors. }
We compare against 4 competitors (NoSplit, @Freeze, @Min, NSG). The first 3 use either no splitting or static splitting strategies and the last one is dynamic. \system{} and all competitors pipeline client computation with server communication and do not reuse the DNN model between iterations. All systems share the same PyTorch code-base.

NoSplit trains solely on the compute tier, streaming as many objects as the training batch size from the \objstore{}. @Freeze always splits at the freezing layer. @Min always splits at the layer with the smallest output (as a tiebreaker the earliest such layer is used) up to and including the freezing layer.

The fourth competitor, which we call NSG, is a dynamic approach based on Neurosurgeon~\cite{kang2017neurosurgeon}. NSG uses the same estimates as \system{} but uses its own splitting algorithm originally designed for inference. That is, it chooses as a split index the one that minimizes the sum of the server-side, client-side and network times. We believe that NSG's splitting algorithm is also a natural approach for TL training because the part of TL before the freeze point (which is also where the splitting occurs) is equivalent to inference (i.e., no backpropagation / only forward pass).

{\it We purposely equip @Freeze, @Min and NSG with \system{}'s batch size adaptation to avoid storage-side OOMs. While batch size adaptation is the contribution of \system{}, lending it to the competitors allows us to analyze the effectiveness of the different approaches to splitting across a wider range of experiments without being impacted by storage-side OOM. NoSplit does not use batch size adaptation as it does not push down.}

\vspace{0.05in}\noindent{\bf Presentation. } Most plots show results for all 7 models and 7 batch sizes using bar charts which contain 7 groups of bars (one group per model) each with 7 bars (one bar for each batch size). For brevity, we use BS to refer to specific batch size values.

\vspace{0.05in}\noindent{\bf Summary of the findings: }
\begin{enumerate}
    \item \system{} improves training runtime compared to all competitors and across all batch sizes (\S\ref{subsec:hapi-vs-nosplit}, \S\ref{subsec:hapi-vs-static-splits}, \S\ref{subsec:hapi-vs-nsg}).
    \item \system{} avoids OOM errors by pushing work to the \objstore{} and using batch size adaptation there (\S\ref{subsec:hapi-vs-nosplit}).
    \item \system{} chooses split points that are close to the optimal points, much more so than the competitors (\S\ref{subsec:optimality}).
    \item \system{}'s benefits remain across a range of network bandwidths between client and storage (\S\ref{subsec:vary-nw-bw}).

\end{enumerate}

\subsection{\system{} vs not splitting (NoSplit)}
\label{subsec:hapi-vs-nosplit}

\begin{figure}[ht]
    \centering
    \includegraphics[width=\linewidth]{figures2024/pdf-hapi-vs-baseline.pdf}
    \caption{Speed-up of HAPI vs NoSplit. The interrupted bars signify infinite speed-up due to OOM in NoSplit.}
    \label{fig:speedup-vs-nosplit}
\end{figure}

Figure~\ref{fig:speedup-vs-nosplit} compares the speed-up of \system{} vs NoSplit across all 7 models and 7 batch sizes. The numbers are averaged over 4 runs for each system. Variability is low, the range for both systems is maximum 1.13 and the coefficient of variation is maximum 0.05. The interrupted bars mean infinite speed-up. That occurs when NoSplit fails with client-side OOM while \system{} successfully completes.

\system{} is always equal or better and the speed-up (when not infinite) is as high as 2.5x. In some cases, \system{} and NoSplit are equal (i.e. AlexNet BS 128, Vgg11 BS 128 - 512, Vgg19 BS 128 - 256). This happens when \system{} realizes that the best approach for that configuration is to not split and thus falls back to behaving like NoSplit. Unfortunately, NoSplit OOMs for larger batch sizes for 6 out of the 7 models while \system{} avoids the OOM. Thus, \system{} enables practitioners, if they wish, to use batch sizes beyond the reach of systems that do not split the DNN.

\subsection{\system{} vs static splitting (@Freeze,@Min)}
\label{subsec:hapi-vs-static-splits}

\begin{figure}[h]
    \centering
    \includegraphics[width=\linewidth]{figures2024/pdf-hapi-vs-freeze.pdf}
    \caption{Speed-up of HAPI over @Freeze. @Freeze always uses the freezing layer as a split point.}
    \label{fig:speedup-vs-freeze}
\end{figure}

\begin{figure}[h]
    \centering
    \includegraphics[width=\linewidth]{figures2024/pdf-hapi-vs-min.pdf}
    \caption{Speed-up of HAPI over @Min. The interrupted bars signify infinite speed-up due to OOM in @Min. @Min always splits at the layer with the smallest output.}
    \label{fig:speedup-vs-min}
\end{figure}

Figures~\ref{fig:speedup-vs-freeze} and \ref{fig:speedup-vs-min} compare the speed-up for \system{} against the static splitting approaches @Freeze and @Min. The results are averaged over at least 4 runs for each system. @Min and @Freeze behave similarly but not identically. The reason is that the layer with the minimum output size is often, but not always, close or identical to the freezing layer. The notable difference is the Transformer model where @Min splits at layer 1 and @Freeze at layer 17. As a result, @Min encounters an OOM on the client side for most batch sizes because it pushes down too little work and thus cannot properly leverage the benefits of batch size adaptation.

The speed-up for AlexNet, ResNet18 and ResNet50 is generally small because the optimal split point is either close or equal to the freeze layer or because several of the best splits yield similar runtimes. However, \system{} shows speed-up for AlexNet BS 128 
because it falls backs to NoSplit while the competitors cannot as they are not dynamic. For the rest of the models the speed-up is larger because the best split points, often chosen by \system{}, are in the middle of the DNN far from the freezing layer. As in~\S\ref{subsec:hapi-vs-nosplit}, for small batch sizes for Vgg11 and Vgg19, \system{} also falls back to NoSplit while the competitors cannot. Even discarding the cases where \system{} falls back to NoSplit, the speed-up is larger than 1.5x for several models. The speed-up generally decreases for larger batch sizes because the optimal split point approaches the freeze index for these configurations. The reason is that as the batch size increases, more and more early splits points become infeasible in order to avoid OOMs on the client side.

\subsection{\system{} vs dynamic splitting (NSG)}
\label{subsec:hapi-vs-nsg}

\begin{figure}[t]
    \centering
    \includegraphics[width=\linewidth]{figures2024/pdf-hapi-vs-nsg.pdf}
    \caption{Speed-up of HAPI over NSG.}
    \label{fig:speedup-vs-nsg}
\end{figure}

Figure~\ref{fig:speedup-vs-nsg} shows the speed-up of \system{} compared to NSG averaged over at least 4 runs each. NSG behaves similarly to @Freeze and @Min because it tends to select splits that come late in the DNN. However, there are differences and NSG improves on @Freeze and @Min (see Vgg19). As before, some of the benefits of \system{} come from its ability to fall back to NoSplit in certain situations but even discarding those cases the speed-up can approach or exceed 1.5x (e.g. Transformer, DenseNet). 

\begin{figure}[h]
    \centering
    \includegraphics[width=\linewidth]{figures2024/pdf-est-acc-opt-hapi-vs-nsg.pdf}
    \caption{(Top) Accuracy of per-epoch runtime estimates for HAPI and NSG correlated with (Bottom) the distance from the optimal split (dots) to the freeze layer (vertical bar).}
    \label{fig:est-acc}
\end{figure}

Figure~\ref{fig:est-acc} illustrates the limitation of NSG's splitting approach. The top part shows the estimation accuracy for \system{} and for NSG. Note that this accuracy is not training accuracy (that remains unchanged in \system{}) but rather per-epoch runtime estimation accuracy which is an important factor in deciding the splitting point. The bottom part of Figure~\ref{fig:est-acc} shows the optimal split point (with bars) and the freeze point (with dots). 

The key insight is that whenever the optimal split occurs significantly before the freezing point (the top of the bar is far from the dots in the  bottom part of Figure~\ref{fig:est-acc}), NSG's estimates are inaccurate because the overlap between the client and server sides is large and NSG does not account for that overlap. As soon as the optimal split is close to the freezing layer (top of the bar is close to the dots) then NSG's estimates are more accurate. Since NSG does not fall back to NoSplit, the bottom part of Figure~\ref{fig:est-acc} only considers the split points greater than 0.

\subsection{Optimality of chosen splits}
\label{subsec:optimality}

Table~\ref{tab:optimality} shows how close to the optimal are the splits chosen by each system across all models and batch sizes. The sub-optimal splits are binned according to how close the resulting runtime is relative to the runtime for the optimal split. There are 4 bins in increments of 5\%. For example, the bin entitled "(10-15]\% off" means that the sub-optimal split chosen yielded a runtime between 10\% and 15\% slower than the optimal split. To obtain the optimal splits we performed a sweep across all split points for each model and batch size.

\begin{table}[h]
    \normalsize
    \def\arraystretch{1.4}
    \centering
    \footnotesize
    \begin{tabular}{|l|l|l|l|l|l|}
        \hline
{\bf \%/MODEL} &  {\bf HAPI}  &  {\bf NSG}   &  {\bf @Freeze}    &  {\bf @Min}   &  {\bf NoSplit}\\\hline
OPTIMAL &  59.2  &  14.8  &  8.2   &  4.1   &  12.2\\\hline
(0-5]\% off &  27.6  &  29.3  &  26.5  &  28.6  &  0\\\hline
(5-10]\% off &  9.2   &  8.5   &  12.2  &  8.2   &  0\\\hline
(10-15]\% off &  2   &  6.4   &  4.1   &  6.1   &  0\\\hline
>15\% off &  2   &  41  &  49  &  40.8  &  34.7\\\hline
OOM &  0   &  0   &  0   &  12.2  &  53.1\\\hline
    \end{tabular}
    \vspace{0.1in}
    \caption{Optimality of chosen splits for \system{} compared to the four competitors.}
    \label{tab:optimality}
\end{table}

\system{} chooses far more optimal splits than the second best system (59\% compared to 14.8\% for NSG). If we consider splits off by at most 5\% from optimal then \system{} is at 86.8\% while NSG reaches 44.1\%. \system{} is not perfect. The main reason for the inaccuracies (especially for the bucket 0-5\% off) is that several split points can yield runtimes very close to optimal and thus small estimation inaccuracies can lead to a sub-optimal choice.

\subsection{The benefits of model reuse}
\label{subsec:model-reuse}

\begin{figure}[h]
    \centering
    \includegraphics[width=\linewidth]{figures2024/pdf-hapi-vs-model-reuse.pdf}
    \caption{Speed-up when reusing the model across requests in \system{} normalized to loading the model for every request.}
    \label{fig:speedup-vs-modelreuse}
\end{figure}

Figure~\ref{fig:speedup-vs-modelreuse} shows the additional speed-up on top of HAPI that is brought by reusing the model already present in the GPU memory in the \objstore{} across different requests, rather than loading the model for every requests. The figure shows averages across at least 4 runs. Sensitivity to model reuse is practically relevant because in a compute-constrained environment like the storage tier it may not be possible to keep all models in memory especially in the practical case when many models are trained simultaneously. This experiment also sheds light on a non-obvious interplay between several factors that influence performance: model execution time, model size and training batch size.

The key insight from this figure is that the benefit of model reuse is dependent on the specific model being used and is correlated with how prominent the model loading overhead is as part of the total processing time spent server-side for a request. When the model loading time constitutes a large portion of the total processing time, eliminating this overhead results in a more significant speed-up. Table~\ref{tab:model-reuse-fwd-pass} shows for \system{} BS 128 the percentage of time spent in loading the model for a request (we discard split 0 because a model is not loaded in that case). AlexNet shows the largest percentage (77\%) and that correlates with the larger speed-up in Figure~\ref{fig:speedup-vs-modelreuse}. Vgg11 and Vgg19 follow with both the percentage in the Table~\ref{tab:model-reuse-fwd-pass} and speed-up in Figure~\ref{fig:speedup-vs-modelreuse}.

A few clarifications are necessary to provide a deeper understanding of the results and their implications. First, the speed-up is not proportional to the percentages in the table. This is due to several factors: (1) a faster server-side computation does not necessarily translate into equal improvements in total runtime (i.e., the client side may take a longer time and thus dictate the overall runtime) and (2) at higher batch sizes concurrency plays an additional role. Second, without the cost of model loading, \system{} may decide on a different split point. For example, for Vgg11 BS 512, \system{} with model reuse splits at 11 while without reuse it falls back to NoSplit. Similarly, for AlexNet BS 128, with model reuse it splits at 16 and without reuse it falls back to NoSplit. Interestingly, in both of these examples, \system{} falls back to NoSplit in order to avoid the impact of loading the model server-side.

\begin{table}[h]
    \large
    \centering
    \def\arraystretch{2.7}
    \resizebox{\linewidth}{!}{%
    \begin{tabular}{|l|c|c|c|c|c|c|c|}
        \hline
        {\bf MODEL} & {\makecell{\bf Alex \\ \vspace{0.2cm} \bf Net}} & {\makecell{\bf Res \\ \vspace{0.2cm} \bf Net18}} & {\makecell{\bf Res \\ \vspace{0.2cm} \bf Net50}} & {\makecell{\bf Dense \\ \vspace{0.2cm} \bf Net}} & {\makecell{\bf Vgg \\ \vspace{0.2cm} \bf 11}} & {\makecell{\bf Vgg \\ \vspace{0.2cm} \bf 19}} & {\makecell{\bf Trans \\ \vspace{0.2cm} \bf former}} \\ \hline
        \makecell{ model \\ \vspace{0.2cm} load} & 0.387 & 0.048 & 0.113 & 0.096 & 0.838 & 0.811 & 0.382 \\ \hline
        \makecell{total \\ \vspace{0.2cm} proc.} & 0.500 & 0.242 & 0.346 & 0.481 & 1.511 & 1.499 & 0.906 \\ \hline
        \makecell{\% \\ \vspace{0.2cm} time} & 77 & 20 & 33 & 20 & 55 & 54 & 42 \\ \hline
    \end{tabular}%
    }
    \vspace{0.1in}
    \caption{The duration and \% of time spent in model loading as part of the total per-request server processing time.}
    \label{tab:model-reuse-fwd-pass}
    \vspace{-7mm}
\end{table}

\subsection{Sensitivity to network bandwidth}
\label{subsec:vary-nw-bw}

\begin{figure}[h]
    \centering
    \includegraphics[width=\linewidth]{figures2024/pdf-nw-vary.pdf}
    \caption{The impact of network bandwidth and \system{}'s dynamicity in choosing a split point.}
    \label{fig:nw-vary}
\end{figure}

Figure~\ref{fig:nw-vary} illustrates at the same time (1) the influence of the network bandwidth between the server and the client and (2) \system{}'s dynamicity in choosing a split point. For this experiment we chose one specific configuration, DenseNet with BS 128. For reference, the figure shows also the performance of NoSplit. 

The key insight is that \system{} adapts the split point to the network bandwidth. It splits at a layer with a larger output size (split 9) for large bandwidth and at a layer with a smaller output size (split 19) when the bandwidth is scarce. Since the network bandwidth is just one factor in \system{}'s decision, the split layer's output size does not strictly decrease with bandwidth. At around 0.5 Gbps, \system{} temporarily falls back to NoSplit as it finds that to be the best option.

The gap between \system{} and NoSplit is largest in the extremes. At low bandwidth, the gap keeps increasing as NoSplit sends more data from the \objstore{} to the client. At high bandwidth, the gap grows initially and then remains constant as the network time becomes insignificant.

\subsection{The benefits of server-side concurrency}

\begin{figure}[h]
    \centering
    \includegraphics[width=\linewidth]{figures2024/pdf-ba-impact.pdf}
    \caption{The benefits of server-side concurrency. Speed-up when using concurrency 2 and 4 normalized to concurrency 1. The x-axis shows 4 chosen splits for each model.}
    \label{fig:ba-impact}
    %\vspace{-5mm}
\end{figure}

%The server-side batch adaptation allows increased concurrency by reducing the amount of GPU memory needed by each request. As a result, more requests can fit and make progress concurrently. Without batch adaptation, for larger batch sizes, only a single requests may fit at a time. After batch adaptation is applied, indeed, the compute capabilities and network bandwidth are also factors but their impact is non-obvious and we believed that this warrants a standalone experiment (Section 5.8 and Figure 14). The severity of how these factors affect the results depends on the model and split point.

\system{} benefits from increased concurrency on the server-side enabled by the combination of lightweight requests and batch size adaptation. The benefits heavily depend on model properties, often in a non-obvious manner. This experiment shows a deeper analysis. Figure~\ref{fig:ba-impact} illustrates the performance gain due to concurrency. To decouple the benefits of concurrency from the splitting algorithm, the figure focuses on server-side speed-up. More precisely, it shows how much faster all requests in an iteration complete when running with a concurrency of 2 or 4 compared to sequentially (concurrency of 1). We consider that all requests complete when the data for the last request in the iteration is received by the client. To obtain a concurrency of 1, 2 and 4 we varied the batch size from 128 to 256 and 512. The grey bars show speed-up with a concurrency of 2 and the colored bars with a concurrency of 4. 

The key insight is that the benefit of concurrency is a function of the split layer output sizes (via the output-related overheads of pickling, network transfers, etc.) relative to the time spent in the forward pass. The output related overheads benefit from parallelization as they are handled by either the CPU or the network. The forward pass is not parallelizable on the Nvidia T4 GPU as it is time-shared. 

In Figure~\ref{fig:ba-impact}, the x-axis shows splits and not batch sizes. We chose 4 splits for each model and ordered them by increasing index for each model. The different colors for concurrency 4 represent different layer output sizes. Where possible, for each model (except Transformer), we selected the first and last split of a particular size. This keeps output-related overheads constant while the forward pass time increases.

The speed-up always drops as the split index grows. The reason for this trend is that, as the split approaches the freeze point, the overheads are low (layer output size is small or minimal) and the forward pass time is largest. Thus, the parallelizable part is smallest. The same effect can be observed by comparing two splits of the same color for a single model. In this case, the overheads are constant (same output size) but the forward pass time grows as the split index increases. 

Across models, the speed-up is greatest for AlexNet since it has the shorter forward pass time. The red color (first two splits for Vgg11 and Vgg19) represents the split point with the largest output size. This explains the increased speed-up for Vgg11 and Vgg19 despite these models not having a small forward pass time.

\section{Related Work}
\label{sec:related-work}

\system{} contains a unique combination of context (\objstore{}), workload (TL), and design decisions (splitting a DNN between cloud tiers, batch size adaptation). While related work exists in these directions, \system{} is the first to combine them in a single cohesive system.
%{\color{red} why works of other applications/domain could not be applied directly to our case?}

\vspace{0.07in}\noindent{\bf Batch size adaptation in ML. }
Dynamic adaptation of the \emph{training} batch size during training was shown~\cite{mascots20-ibm, iclr2018-GoogleIncreaseBatch,kdd2017-batch} to reduce training times without loss in accuracy. These ideas are complementary to \system{}'s server-side batch size adaptation because they can be applied in \system{}'s training phase on the client side. The key insight in \system{} is that during the same iteration, it is useful to configure different batch sizes for feature extraction compared to training.

\vspace{0.07in}\noindent{\bf Splitting ML compute between clouds/tiers.  }
Neurosurgeon~\cite{kang2017neurosurgeon} splits ML \emph{inference} between the cloud and edge devices to achieve low latency, low energy consumption, and high data center throughput. The partitioning is automatic and adapts to dynamics in server load and network bandwidth. Compared to Neurosurgeon, besides considering a more complex application (i.e., TL), \system{} dynamically manages storage-side concurrency via batch size adaptation. NDPipe~\cite{ndpipe-asplos24} is a parallel effort to ours focusing on accelerating training and inference for image data by using near-data processing techniques to photo storage servers equipped with inexpensive commodity GPUs. \system{} differs in several technical aspects, notably by analyzing concurrency in the \objstore{}, proposing storage-side batch size adaptation, avoiding OOMs and by fine-tuning without impacting the accuracy. 

\vspace{0.07in}\noindent{\bf Offloading compute to storage. }
Work on pushdowns to \objstore{} focuses on leveraging the limited subset of SQL supported by Amazon S3 Select-like systems. PushdownDB~\cite{icde2020-pushdowndb} analyzes which DBMS primitives can be cost-effectively moved into S3 and how more complex operations (e.g., joins) can be re-implemented on top of S3. FlexPushdownDB~\cite{vldb2021-flexpushdowndb} introduces separable operators that combine data from caches with results from pushdowns. Other work analyzes pushing down computation to storage tiers not limited by S3 Select-like APIs. Rhea~\cite{nsdi2013-rhea} uses static analysis of Java bytecode to generate storage-side filters for Hadoop applications. Rhea also mitigates the network bottleneck to storage but targets a different workload, does not split an application and is not concerned with OOMs. NeSSA~\cite{10.1145/3599691.3603404} filters training data in FPGA-based SmartSSDs, transferring out relevant data subsets that result in comparable training accuracy to the full dataset. This technique is complementary to \system{} which trains on the entire dataset. DDS~\cite{10.14778/3681954.3682002} offloads DBMS read operations to DPUs, reducing latency and CPU usage on storage servers.

\vspace{0.07in}\noindent{\bf Mitigating I/O bottlenecks. }
Recent works such as Fluid~\cite{10249214} and SiloD~\cite{10.1145/3552326.3567499} focus on reducing I/O bottlenecks in cloud-based AI training. They emphasize on efficient management of storage and compute resources and on improving data access latency through advanced caching and scheduling techniques. In contrast, HAPI addresses the I/O bottleneck using a different approach by directly minimizing data movement by offloading computation to the storage layer.

%co-designs the cluster scheduler and the cache subsystems

%\textcolor{blue}{DeepPlan\cite{deepplan-eurosys23} minimizes deep learning inference latency by optimizing model provisioning from host to GPU memory. Traditional methods rely on caching models in GPU memory to avoid delays, which often leads to high operational costs due to the need for over-provisioning GPU resources. It uses direct-host-access (DHA) to allow GPUs to access main memory directly, reducing cold-start times by selectively loading only necessary layers. Additionally, it employs parallel model transmission across multiple GPUs to speed up model loading by dividing workloads. These techniques make it effective in reducing inference delay and enhancing performance in high-concurrency environments. HAPI's model reuse shares the same goal as DeepPlan in minimizing latency by reducing the overhead of model loading.}

%references about how to solve memory problems -- we do batch adaptation for that
%\item \url{https://www.usenix.org/conference/fast21/presentation/bae} -- interesting -- uses SSD in DNN training -- another way to solve the memory problems (they say one solution is to use host memory -- for offloading) yet, this paper uses fast SSD to solve the GPU memory problem. It depends on a direct very fast links between SSD and GPU though. This can be used as a reference to the GPU memory problem (see Fig. 2 in the paper).
%there should be also other papers that use the host memory and pipelining to solve this problem -- we can check that later.

\vspace{0.07in}\noindent{\bf Model, pipeline and tensor parallelism.}
Splitting in \system{} resembles model~\cite{dehghani2023scalingvisiontransformers22}, pipeline~\cite{10.1145/3673038.3673047, pipedream-sosp19,gpipe-neurips19}, and tensor~\cite{tenplex} parallelism, which aim to distribute the training of large models across GPU clusters. Recent systems like Aceso~\cite{10.1145/3627703.3629554}, AutoPipe~\cite{10.1145/3673038.3673047} and Tenplex~\cite{tenplex} further advance these strategies. Meta's Glow compiler~\cite{rotem2019glowgraphloweringcompiler} partitions models across accelerators while applying compiler optimizations and code generation techniques. \system{} differs from these parallelism approaches in several key ways. Its primary objective is to mitigate the COS network bottleneck by splitting TL's feature extraction phase. The concerns differ as \system{} handles multiple clients concurrently  storage-side using batch size adaptation, and there is no backpropagation between \system{}'s client and server.

%Recent systems like Aceso~\cite{10.1145/3627703.3629554}, which provides an iterative framework to alleviate bottlenecks in large-scale DNN training, AutoPipe~\cite{10.1145/3673038.3673047}, which dynamically adjusts pipeline parallelism using reinforcement learning, and Tenplex~\cite{tenplex}, which supports dynamic parallelism changes for elasticity, redeployment, or failures, further advance these strategies. 

\vspace{0.07in}\noindent{\bf Data ingestion pipelines.}
The feature extraction phase in \system{} resembles data ingestion/pre-processing pipelines~\cite{tf.data-vldb2021, data-ingestion-pipelines-isca22-meta-kozyrakis, preprocessing-pipelines-sigmod22, pecan-atc24, sophon-hotstorage24} because of their goal (data preparation) and the critical focus on efficiently delivering data for training. However, these pipelines are far more general in nature (e.g., can even run user-defined functions~\cite{tf.data-vldb2021}) and can be very large (e.g., serve data at TB/s~\cite{data-ingestion-pipelines-isca22-meta-kozyrakis}), which makes them challenging to run entirely inside the \objstore{}. SOPHON~\cite{sophon-hotstorage24} offloads specific pre-processing tasks to remote storage. In \system{}, because the feature extraction phase is part of the DNN, it needs to be optimized as part of the training job with fewer knobs compared to full-fledged pipelines.

\vspace{0.07in}\noindent{\bf Related optimizations.}
DeepPlan~\cite{deepplan-eurosys23} and PipeSwitch~\cite{pipeswitch-osdi20} are complementary works to \system{} that optimize model loading from host to GPU memory and can be applied to HAPI servers when model reuse is not possible. Intel’s OpenVINO Toolkit~\cite{OpenVINO9537452} optimizes inference by applying hardware-specific optimizations. OpenVINO can accelerate and improve the efficiency of HAPI's feature extraction phase.

\section{Discussion and future Work}

\vspace{0.07in}\noindent{\bf Impact of features/limitations of \objstore{} platforms.}
Certain features and limitations of \objstore{} platforms can facilitate or hinder \system{}'s performance and implementation. The faster the storage-side GPU is, the better the runtime of the pushdown becomes. A fast storage medium for TL inputs can mitigate potential storage bottlenecks when feeding those inputs to the DNN pushdown and thus the storage-side GPU bandwidth can be efficiently utilized. Though not always available in the \objstore{}, storage and network-level bandwidth isolation between pushdowns and regular data transfers (i.e., transfers out of the \objstore{} that do not involve pushdowns) would help increase the predictability of pushdowns.

\vspace{0.07in}\noindent{\bf Automating model splitting.}
In future work, we aim to automate the model customization process for DNNs to enable them to run the forward pass between arbitrary start and end layers, allowing computation to be split at any chosen layer. This involves automatically analyzing the model's computational graph to detect layers involved in non-sequential dependencies, such as skip connections or residuals, ensuring safe model partitioning without manual intervention. In PyTorch, this can be achieved by leveraging tools like torch.fx to trace the graph and identify dependencies. 
%By integrating this automation, we aim to simplify the usage of HAPI, making it more accessible and facilitating its adoption.

\vspace{0.07in}\noindent{\bf Efficient parameter fine-tuning methods.}
While our work focuses on traditional fine-tuning methods in TL, recent advances in efficient parameter fine-tuning, such as low-rank adaptation (LoRA)~\cite{hu2021loralowrankadaptationlarge}, and quantization techniques~\cite{li2024loftq,10.5555/3666122.3666563} have shown significant promise in reducing the number of trainable parameters and computational requirements while preserving model performance. These techniques are particularly useful for large-scale models and resource-constrained environments. Given that, for the moment, \system{} only considers freezing early layers and limiting backpropagation to later stages, adapting these methods is not straightforward. We plan to investigate how such approaches could be integrated into a storage-disaggregated architecture.

\section{Concluding Remarks}

%\vspace{0.05in}\noindent{\bf Summary.} 
\system{} is a transfer learning (TL) fine-tuning system that spans the compute and \objstore{} tiers and introduces two new techniques that address challenges introduced by storage disaggregation. \system{} judiciously splits the DNN between tiers by pushing down to storage part of TL's feature extraction phase. This mitigates the cloud network bottleneck and also reduces training time by carefully balancing processing time across tiers. Crucially, splitting also decouples the batch size used during the training and feature extraction phases, facilitating the use of batch size adaptation in the \objstore{} without affecting the overall training accuracy. This enables increased pushdown concurrency by greatly reducing the amount of memory necessary for the pushdowns. \system{} yields up to 2.5× training speed-up while choosing in 86.8\% of cases the best performing split point or one that is at most 5\% off from the best.

%\system{} shows up to 2.5x speed-up in application runtime and chooses a split point that %is either optimal or at most 5\% off in performance in 86.8\% of the cases.

%%
\begin{comment}
\vspace{0.05in}\noindent{\bf Beyond TL.}  While our techniques can be generalized to other data science pipelines, TL's properties made it a perfect fit for splitting. Specifically, two aspects of TL's feature extraction are very useful for splitting: (1) the absence of backpropagation and (2) the reduction of data size. The absence of backpropagation from the \objstore{} allows the server to be lightweight and stateless and avoids additional network load. The reduction in data size helps reduce network transfers, reducing consequently the training time. Finding an application with similar characteristics to benefit from splitting is an avenue for future work.
\end{comment}

%In this paper, we use \system{} to split TL computation among storage and compute.

%\vspace{0.05in}\noindent{\bf Future work.} As future work, we consider caching the output of feature extraction across epochs since the feature extraction computation is deterministic. This, however, implies a trade-off between storing intermediate data versus performing redundant computation, which is outside the scope of this paper and depends on how well storage servers are provisioned. We are also considering re-computing the splitting point at finer-granularity (one or several batches) to account for variation in resource availability.

\section*{Acknowledgments}
We thank our shepherd, Pankaj Mehra, and the anonymous reviewers for their insightful comments. This work has been partially funded by a research grant from the Huawei Munich Research Center, Germany.

\bibliographystyle{ACM-Reference-Format}
%\bibliography{references}

\end{document}